\documentclass{article}
\usepackage{spconf}
\usepackage{times}
\usepackage{latexsym}
\usepackage{amsmath}
\usepackage{amssymb}
\usepackage{graphicx}
\usepackage{makecell}
\usepackage[table]{xcolor}
\usepackage{colortbl}
\usepackage[hidelinks]{hyperref}
\usepackage{svg}
\graphicspath{{./assets/figs/}}
\usepackage{booktabs}
\usepackage{multirow}
\usepackage{xcolor}
\usepackage{colortbl}
\usepackage{soulutf8}
\usepackage{xcolor}

\sethlcolor{green}   
\newcommand{\hls}[1]{\sethlcolor{green}\hl{#1}}   
\newcommand{\hld}[1]{\sethlcolor{pink}\hl{#1}}     
\newcommand{\hlo}[1]{\sethlcolor{cyan}\hl{#1}}     

\title{Stuttering-Aware Automatic Speech Recognition \linebreak for Indonesian Language}
\name{
    \begin{tabular}{c}
    Fadhil Muhammad\textsuperscript{1,*},
    Alwin Djuliansah\textsuperscript{1,*},
    Adrian Aryaputra Hamzah\textsuperscript{1,*},
    Kurniawati Azizah\textsuperscript{1}
    \end{tabular}
    \thanks{* Equal Contribution}
}

\address{
    \textsuperscript{1}Faculty of Computer Science, Universitas Indonesia \\
    \small \texttt{\{fadhil.muhammad23, alwin.djuliansah, adrian.aryaputra\}@ui.ac.id} \\
    \small \texttt{kurniawati.azizah@cs.ui.ac.id}
}

\begin{document}
\maketitle

\begin{abstract}
Automatic speech recognition systems have achieved remarkable performance on fluent speech but continue to degrade significantly when processing stuttered speech, a limitation that is particularly acute for low-resource languages like Indonesian where specialized datasets are virtually non-existent. To overcome this scarcity, we propose a data augmentation framework that generates synthetic stuttered audio by injecting repetitions and prolongations into fluent text through a combination of rule-based transformations and large language models followed by text-to-speech synthesis. We apply this synthetic data to fine-tune a pre-trained Indonesian Whisper model using transfer learning, enabling the architecture to adapt to dysfluent acoustic patterns without requiring large-scale real-world recordings. Our experiments demonstrate that this targeted synthetic exposure consistently reduces recognition errors on stuttered speech while maintaining performance on fluent segments, validating the utility of synthetic data pipelines for developing more inclusive speech technologies in under-represented languages.
\end{abstract}



\section{Introduction}
Automatic Speech Recognition (ASR) has become a central component of many modern applications, ranging from virtual assistants to transcription services and accessibility tools. Although recent ASR systems have achieved strong performance on fluent speech, their robustness to speech disorders remains limited \cite{huang:stuttered-asr}. Stuttering is one example of a condition that continues to challenge current models.

The impact of this mismatch is evident in practical use. People who stutter frequently encounter recognition errors, interrupted utterances caused by premature end-of-speech detection, and, in some cases, incorrect words generated as the system attempts to interpret long pauses \cite{huang:stuttered-asr, bleakley:stammering-ux}. These issues reduce the usability of voice-based interfaces and highlight the need for ASR models that explicitly account for stuttering behavior.

For Indonesian, the problem is further compounded by the lack of dedicated resources. While Indonesian is widely spoken, existing speech datasets mainly contain fluent, read speech \cite{adila:indo-asr}  and do not represent the characteristics of stuttered utterances. As a result, current ASR models trained on Indonesian data often struggle to generalize to users with speech disfluencies.

In addition to this data gap, several linguistic properties of Indonesian introduce their own challenges. The language makes extensive use of reduplication as a regular morphological process, which can resemble stutter-like repetition at the surface level. For example, distinguishing between a stuttered form such as “sa-sa-sapi” (disfluent version of sapi) and the valid reduplicated word “sapi-sapi” (plural form) requires sensitivity to prosodic cues rather than simple text-pattern matching. These aspects suggest that addressing stuttering in Indonesian ASR requires both appropriate data and methods tailored to the language’s structure.

\section{Related Work}
To situate this research within existing literature, this section reviews prior work on ASR robustness, stuttering and disfluency analysis, and linguistic features of Indonesian that pose challenges for stuttering-aware recognition.

\subsection{ASR Limitations for Disfluent Speech}
Automatic Speech Recognition (ASR) systems have achieved strong performance on fluent speech, but their accuracy drops sharply when faced with disfluent or stuttered utterances. Disfluencies such as repetitions, prolongations, interjections, and blocks introduce acoustic and temporal irregularities that conventional models fail to capture. As a result, recognition errors, truncated utterances, and misinterpretations are common when users with stuttering attempt to interact with voice‑based systems.

A research from \cite{huang:stuttered-asr} showed that conventional ASR systems often treat disfluencies as irregular noise, leading to systematic recognition failures in spontaneous speech. Furthermore, research from \cite{bleakley:stammering-ux, wu:videoconf-stutter} demonstrated that mainstream ASR systems misinterpret stuttered utterances, reducing usability for individuals with speech disorders.

Together, these findings establish that while ASR has matured in handling fluent and noisy speech, it remains fragile when faced with pathological disfluencies such as stuttering. This limitation highlights the need for models explicitly designed to accommodate diverse speech patterns.


\subsection{Stuttering-Aware ASR}
While ASR systems struggle with stuttered speech, research has produced methods to detect stuttering events directly from acoustic signals. Deep learning approaches, such as those introduced by \cite{kourkounakis:residual-disfluency}, can identify repetitions, prolongations, and blocks with high accuracy. Other studies, including \cite{amann:disfluency-asr, mujtaba:interspeech2024inclusiveasr} combine disfluency detection with augmentation strategies to improve recognition robustness. These methods demonstrate that stuttering detection is feasible and can serve as a foundation for more inclusive ASR.

However, such detection techniques remain largely confined to English datasets and have not been integrated into Indonesian ASR pipelines. Existing Indonesian speech corpora focus on fluent, read speech and lack annotations for disfluencies. As a result, while detection methods exist in the broader ASR research community, Indonesian systems continue to operate without mechanisms to identify or adapt to stuttering behavior. This gap reinforces the need for localized research that bridges detection methods with Indonesian language resources.

\subsection{Limited Indonesian Stuttering Resources}

The development of robust ASR systems for stuttered speech depends heavily on the availability of representative datasets. At present, most stuttering corpora are designed for English, reflecting the dominance of English in speech technology research. Resources such as the UCLASS Stuttering Archive \cite{howell:uclass} and the SEP‑28k dataset \cite{lea:sep28k} provide valuable material for training and evaluation, but they are limited to English speakers. Recent shared tasks, including the IEEE SLT 2024 StutteringSpeech Challenge \cite{amann:disfluency-asr}, have further advanced benchmarking, yet again within the English language domain.

For Indonesian, however, no equivalent stuttering corpus exists. Existing Indonesian speech datasets such as Indonesian Speech Recognition Corpora presented at IALP 2024 \cite{adila:indo-asr} primarily consist of fluent, read speech and do not capture the characteristics of disfluent utterances. This absence of stuttering‑specific resources means that ASR models trained on Indonesian data struggle to generalize to users with speech disorders, leaving a significant accessibility gap.

\section{Methodology}
This section outlines the methodological framework used to develop a stuttering-aware ASR system for Indonesian. We describe the process of constructing a synthetic Indonesian stuttering dataset, the stuttering patterns modeled, and the text-to-speech (TTS) pipeline used to generate training audio. We then detail the ASR model, training procedures, and evaluation setup.

\subsection{Dataset}
We constructed a stuttering-augmented dataset derived from fluent Indonesian text. The objective was to simulate dysfluency patterns commonly described in \cite{darmadie:indonesian-stuttering} which includes repetitions (e.g., “a-a-aku”, “sa-sa-saya”), prolongations (e.g., “sssaya”), and interjections (e.g., “saya mau umm makan”) so that the ASR model is exposed to stutter-like acoustic variations during training.


\subsubsection{Source Corpus}
As the basis for our stuttering-augmented dataset, we utilized the Indonesian utterance portion of the Mozilla Common Voice corpus \cite{commonvoice:23}. Common Voice is a large-scale, open-source collection of speech recordings contributed by volunteers, designed to support inclusive and multilingual ASR research. The Indonesian subset primarily consists of read, fluent speech sampled from diverse speakers across regions and demographics. While the corpus provides valuable coverage of standard Indonesian pronunciation, it does not contain stuttered or disfluent utterances. This characteristic makes it suitable as a clean baseline from which we can systematically introduce synthetic stuttering patterns, ensuring that the augmented dataset reflects both natural Indonesian speech and the dysfluency phenomena described in \cite{darmadie:indonesian-stuttering}.

\subsubsection{Stutter Types}
For this study, we focus on three primary categories of stuttering phenomena. 
\textbf{Repetition} refers to the reiteration of phonemes, syllables, words, or phrases (e.g., “sa-sa-saya lapar”). 
\textbf{Prolongation} involves the lengthening of an onset consonant or vowel within a word (e.g., “sss-saya sudah makan”). 
\textbf{Interjection} denotes the insertion of filler expressions such as \textit{“emm”, “anu”, “hmm”} within otherwise fluent speech (e.g., “saya emm tadi makan”).

\subsubsection{Stutter Text Generation}
To generate a pseudo-stutter dataset, we apply two complementary text-augmentation strategies to the original Indonesian transcripts. The first strategy is a deterministic algorithm inspired by previous works \cite{mujtaba:interspeech2024inclusiveasr} that inserts controlled disfluencies based on predefined probabilities for repetition, prolongation, and interjection. This approach allows us to model specific stuttering behaviors consistently and to enforce linguistic constraints, such as preventing interjections from appearing at the end of a sentence and ensuring that repetitions or prolongations occur only in phonologically plausible positions. The algorithm also uses context-sensitive rules to select appropriate interjections, distinguishing between short fillers (e.g., “emm”, “hmm”) and longer “thinking” expressions (e.g., “apa ya…?”, “sebentar…”).

The second strategy relies on a large language model prompted to rewrite fluent Indonesian text into stutter-like variants. This enables the generation of more naturalistic and contextually grounded disfluencies that reflect patterns not easily captured by hand-crafted rules, such as subtle phrase-level repetitions or hesitations shaped by semantic context.

The deterministic algorithm is implemented as a probabilistic text transformation pipeline. For each word in the input sentence, the system decides (according to a global disfluency probability) whether to introduce a modification. Repetition is implemented by duplicating either the initial syllable (e.g., “sa-sa-saya”) or the entire word (“mau mau mau”), whereas prolongation extends the initial or final consonant (e.g., “sssaya”). Interjections are inserted strictly between word boundaries and are chosen using a small context model that accounts for sentence position and discourse markers. By combining controlled rule-based operations with a larger set of variable, context-aware outputs from an LLM, the final pseudo-stutter dataset captures both predictable and naturalistic features of Indonesian stuttering.

\subsubsection{Stutter Speech Generation}
After generating the pseudo-stuttered text, it is converted into audio using OpenAI’s text-to-speech system with \emph{gpt-tts-4o-mini}. A system prompt is added to instruct the model to simulate stuttering based on the already transformed text. Speaker selection is sampled uniformly at random from all available voices, and the speed is sampled uniformly from the range $[0.75, 1.25]$. For both clean and stuttered text, we generate corresponding audio to ensure a direct and consistent comparison between fluent and disfluent speech set.



\subsubsection{Data Overview}

The final dataset consists of four components derived from the synthetic augmentation process, enabling direct comparison between fluent and stuttered speech.

\begin{itemize}
    \item \textbf{Original Text}: Fluent Indonesian transcripts used as ground-truth labels and as the source for generating stuttered variants.
    \item \textbf{Original Audio}: Clean speech recordings corresponding to the original text.
    \item \textbf{Stuttered Text}: Augmented transcripts containing repetitions, prolongations, and interjections, generated using rule-based transformations and a language model.
    \item \textbf{Stuttered Audio}: Synthetic speech produced from the stuttered text using a text-to-speech system.
\end{itemize}

\begin{table*}[ht!]
\centering

\renewcommand{\arraystretch}{1.3} 
\setlength{\tabcolsep}{8pt}       

\begin{tabular}{llcccc}
\toprule
& & \multicolumn{2}{c}{\textbf{Dev Set}} & \multicolumn{2}{c}{\textbf{Test Set}} \\
\cmidrule(lr){3-4} \cmidrule(lr){5-6}
\textbf{Model / Strategy} & \textbf{Test Condition} & \textbf{WER} & \textbf{CER} & \textbf{WER} & \textbf{CER} \\
\midrule

\rowcolor{gray!10} 
\multicolumn{6}{l}{\textit{Baseline (Zero-shot)}} \\
\multirow{2}{*}{Whisper Base} 
  & Stutter & \textendash & \textendash & 0.818 & 0.460 \\
  & Clean   & \textendash & \textendash & 0.100 & 0.022 \\
\addlinespace 

\rowcolor{gray!10}
\multicolumn{6}{l}{\textit{Fine-tuned Models}} \\
\multirow{2}{*}{\textbf{Stutter only}}
  & Stutter
  & \textbf{0.101} & \textbf{0.039}
  & \textbf{0.126} & \textbf{0.055} \\
  & Clean
  & \textbf{0.054} & \textbf{0.014}
  & \textbf{0.064} & \textbf{0.016} \\
\addlinespace

\multirow{2}{*}{Stutter + Clean}
  & Stutter
  & 0.154 & 0.066
  & 0.181 & 0.084 \\
  & Clean
  & 0.107 & 0.040
  & 0.076 & 0.019 \\
\bottomrule
\end{tabular}

\caption{Comparison of ASR performance. The baseline Zero-shot model is compared against fine-tuning strategies on stuttered and clean test sets. Values are reported in WER and CER (lower is better). Best results are highlighted in \textbf{bold}.}
\label{tab:results}
\end{table*}

\begin{figure}[ht]
    \centering
    \includegraphics[width=\linewidth]{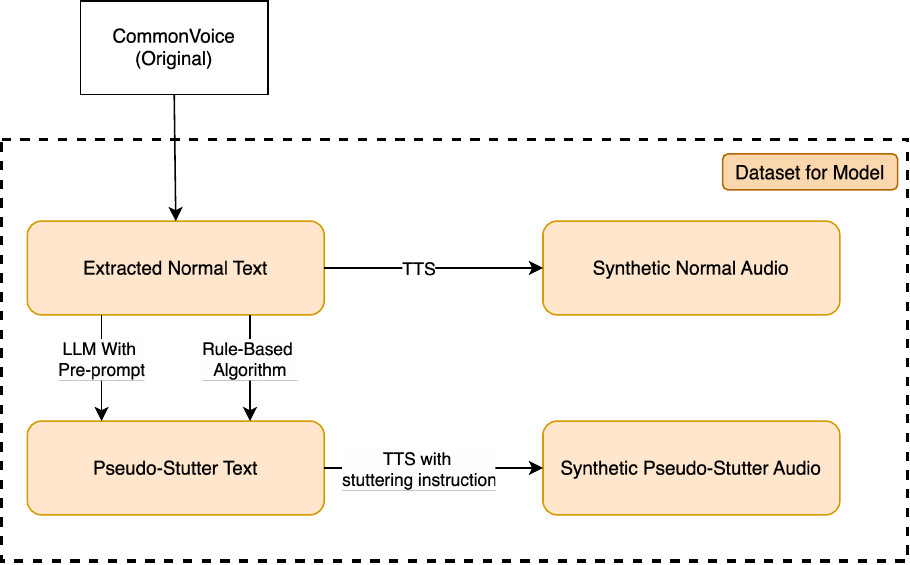}
    \caption{Dataset generation workflow using rule-based and LLM approaches, followed by audio generation using text-to-speech model}
    \label{fig:dataset-flow}
\end{figure}

\subsection{Model}

We adopt Whisper, a sequence-to-sequence automatic speech recognition (ASR) model based on an encoder--decoder Transformer architecture. Unlike CTC-based approaches, Whisper directly models the conditional probability of an output token sequence given an audio input. This formulation allows the model to handle long-range dependencies and flexible temporal variations in speech, which are common in disfluent utterances such as repetitions, prolongations, and interjections.

\subsection{Preprocessing}

All data undergo a consistent preprocessing pipeline to ensure compatibility with the Whisper ASR model and to reduce sources of variability unrelated to speech content. Preprocessing is applied uniformly to both fluent and stuttered data to maintain comparability across experimental conditions.

For textual data, we perform standard normalization steps to minimize orthographic variation. These steps include converting all text to lowercase, removing punctuation marks, collapsing consecutive whitespace characters into a single space, and trimming leading and trailing whitespace. This normalization helps stabilize training and evaluation by ensuring that differences in model performance are attributable to acoustic and temporal characteristics rather than superficial textual inconsistencies.

For audio data, all recordings are resampled to a sampling rate of $16$~kHz using the \textit{librosa} library and padded to $30$ seconds, matching the expected input format of the Whisper model. This resampling step ensures a uniform temporal resolution across all audio samples, including both original recordings and synthetically generated speech. No additional signal-level transformations are applied, allowing the model to learn directly from the raw acoustic representations.

\section{Experiments}
In this section, we present the experimental framework used to assess the effectiveness of synthetic stuttering augmentation for Indonesian speech recognition. We detail the training configurations and evaluation protocol, and report results on both stuttered and clean speech conditions.

\subsection{Experimental Setup}
We establish our baseline using a Whisper-small model fine-tuned on clean Indonesian speech, initialized from the publicly available \texttt{cahya/whisper-small-id} checkpoint. This model serves as a conventional ASR baseline that has not been adapted to disfluent speech.

To evaluate the impact of synthetic stuttering augmentation, we fine-tune the same Whisper model under two additional training configurations: (1) training exclusively on synthetic stuttered speech and (2) training on a mixture of synthetic stuttered and clean fluent speech. All models share the same architecture and training procedure, differing only in the composition of the fine-tuning data. This controlled setup allows us to directly assess the contribution of synthetic stuttering data to recognition performance on stuttered Indonesian speech.

The dataset comprised synthetic utterances (both fluent and stuttered) generated via text-to-speech systems. We employed a speaker-stratified split to ensure that the training, development, and test sets are disjoint in terms of speakers. For both clean and stuttered audio, the dataset is divided into 9,928 utterances for training, 1,102 for development, and 1,103 for testing. Training was conducted on a Nvidia P100 GPU with a batch size of 16 for training, utilizing the AdamW optimizer with a learning rate of $\alpha = 1 \times 10^{-5}$ and a Linear scheduler ($lr_{\min} = 1 \times 10^{-8}$).

\subsection{Evaluation Metrics}

For the Automatic Speech Recognition (ASR) task, we report the Word Error Rate (WER) and Character Error Rate (CER). These metrics quantify the edit distance between the predicted transcription and the ground truth:
\begin{equation}
\text{Error Rate} = \frac{S + D + I}{N}
\end{equation}
where $S$, $D$, and $I$ denote substitutions, deletions, and insertions, and $N$ represents the total count of reference units.

\subsection{Experiment Results}

The results summarized in Table~\ref{tab:results} indicate that fine-tuning Whisper on synthetic stuttered speech leads to clear and consistent gains over the zero-shot baseline, particularly when evaluated on stuttered input. Across both the development and test sets, the model trained exclusively on stuttered speech achieves the lowest word and character error rates.

On the stuttered test condition, stutter-only fine-tuning yields a substantial reduction in both WER and CER compared to joint fine-tuning with clean data. This finding suggests that focused exposure to disfluent speech allows the model to better internalize the acoustic and temporal characteristics associated with stuttering. Notably, this specialization does not appear to harm recognition of fluent speech. On the contrary, the stutter-only model also attains the lowest error rates on clean speech, indicating that adaptation to disfluencies can generalize beyond the specific training condition.

In contrast, jointly fine-tuning on stuttered and clean speech results in inferior performance on stuttered input. A plausible explanation is that simple data mixing may reduce the effective emphasis on stuttering patterns during training, thereby limiting the model’s ability to fully adapt to disfluent speech. While further investigation is needed to better understand this behavior, the observed results underscore the importance of targeted data augmentation strategies when adapting ASR systems to atypical speech patterns.

\subsection{Error Analysis on Stuttered Speech}

To analyze the behavior of the best-performing model on stuttered speech, we examine representative transcription errors and highlight the affected spans for clarity. Different highlight colors are used to indicate distinct error types.

\subsubsection{Phonetic Substitution with Semantic Shift}
\begin{quote}
\small
Reference: TERUS KENAPA? \\
Prediction: TERUS \hls{KEREN APA}?
\end{quote}

In this example, a phonetic substitution occurs in which \textit{KENAPA} is replaced by the acoustically similar phrase \textit{KEREN APA}. Although plausible at the acoustic level, this substitution alters the intended meaning of the utterance.

\subsubsection{Lexical Distortion under Repetition}
\begin{quote}
\small
Reference: SAYA BERANDAI-ANDAI APAKAH TOM MASIH MENGINGAT BAGAIMANA CARA UNTUK MELAKUKANNYA. \\
Prediction: SAYA \hls{MERANDANG-ANDANG} APAKAH TOM MASIH MENGINGAT BAGAIMANA CARA UNTUK MELAKUKANNYA.
\end{quote}

Here, repetition within a lexical unit leads to distortion, producing a non-standard form. Despite this error, the surrounding context is correctly transcribed, indicating robustness at the sentence level.

\subsubsection{Deletion and Phonetic Substitution}
\begin{quote}
\small
Reference: \hld{RASA} SALING MENGHORMATI DI ANTARA MEREKA MUNCUL DARI PERISTIWA TERSEBUT. \\
Prediction: SALING MENGHORMATI DI ANTARA MEREKA MUNCUL DARI \hls{PASTIWA} TERSEBUT.
\end{quote}

This example illustrates a deletion of a content word and a phonetic substitution. While transcription fidelity is reduced, the core semantic meaning of the sentence remains largely intact.

\subsubsection{Plural/Singular Ambiguation}
\begin{quote}
\small
Reference: SEKARANG KOTA INI TERLETAK DI PROVINSI UTRECHT. \\
Prediction: SEKARANG \hlo{KOTA-KOTA} INI TERLETAK DI PROVINSI UTRECHT.
\end{quote}

In this case, a singular noun is incorrectly rendered as plural. This form of plural/singular ambiguation suggests uncertainty in number marking, potentially influenced by rhythmic or prosodic irregularities associated with stuttered speech. While the sentence remains syntactically well-formed, the predicted output introduces a change in meaning that is not supported by the reference.

\subsubsection{Summary}

Across these examples, errors predominantly arise from phonetic substitutions, lexical instability under repetition, deletions, and ambiguation in number marking. Importantly, many errors preserve overall sentence structure and partial semantic content, indicating that the model captures global context even when local transcription inaccuracies occur.

\section{Conclusion}

In this work, we investigated the use of synthetic stuttering augmentation to improve automatic speech recognition for Indonesian, a low-resource setting in which publicly available stuttered speech data is largely absent. By introducing repetitions, prolongations, and interjections into fluent text and converting the augmented text into speech, we constructed a synthetic dataset that captures key surface-level characteristics of stuttered speech. This dataset was used to fine-tune a pre-trained Whisper model within a sequence-to-sequence framework.

Our experimental results demonstrate that targeted fine-tuning on synthetic stuttered speech substantially improves recognition performance on disfluent input. Notably, fine-tuning exclusively on stuttered speech consistently outperformed joint fine-tuning with clean data, not only on stuttered test sets but also on clean speech. This finding suggests that focused adaptation to disfluencies can enhance robustness without sacrificing generalization to fluent speech. Moreover, the strong degradation observed in the zero-shot baseline highlights the limitations of conventional ASR models when confronted with atypical speech patterns.

Overall, this study shows that even relatively simple synthetic augmentation strategies can be effective for adapting modern sequence-to-sequence ASR models to stuttered speech. More broadly, it underscores the importance of addressing speech variability in ASR research and contributes toward the development of more inclusive speech technologies for underrepresented speaker populations.

\section{Future Work}

Several directions for future research emerge from this work. First, while the synthetic stuttering generation employed in this study captures common disfluency patterns, it remains a simplification of real stuttered speech. Future efforts should focus on validating and extending these findings using real-world stuttered Indonesian speech, as well as incorporating more nuanced phenomena such as blocks, variable severity levels, and speaker-dependent stuttering behaviors.

Second, the observed performance gap between stutter-only and mixed fine-tuning suggests that more principled data balancing or curriculum learning strategies may be required when combining fluent and disfluent speech. Exploring adaptive sampling, loss reweighting, or staged fine-tuning could provide deeper insight into how ASR models internalize disfluency-related cues.

Third, this work focuses on transcription accuracy as the primary evaluation metric. Future studies may consider complementary objectives, such as explicitly modeling or preserving disfluencies in the output, depending on downstream use cases. Additionally, integrating prosodic or acoustic features beyond raw waveform input may further improve robustness to irregular speech timing.

Finally, extending this approach to other languages and speech disorders would help assess the generality of synthetic disfluency augmentation. Such investigations would contribute to a broader understanding of how modern ASR systems can be adapted to better reflect the diversity of human speech.

\section{Limitations}

Our data pipeline relies too heavily on synthetic data generation, and there is currently no validation stage to ensure the quality of the dataset. There remains a gap that must be addressed to ensure the dataset aligns with characteristics of Indonesian stuttering as it occurs in natural, everyday language use. It is also necessary to ensure that models fine-tuned on our dataset can generalize to real-world scenarios. However, our current work addresses the lack of stuttering data in Indonesian language by introducing an automatic method for generating stuttered Indonesian speech.

\section{Ethics Statement}

We use utterances from the Common Voice dataset, which is released under a public-domain–compatible CC0 license, allowing reuse for research purposes. All data are processed in accordance with the dataset’s licensing terms. Synthetic speech is generated using text-to-speech models to address data availability and the scarcity of Indonesian stuttering speech resources. We acknowledge the potential risk of bias introduced by the use of TTS systems. The generated audio is used strictly for research and model development purposes.


\bibliographystyle{IEEEbib}

\end{document}